\documentclass[letterpaper]{article}

\usepackage{natbib,alifeconf}  
\usepackage{hyperref}

\title{Self-Replication, Spontaneous Mutations, and Exponential Genetic Drift in Neural Cellular Automata}
\author{Lana Sinapayen \\
\mbox{}\\
Sony Computer Science Laboratories, Kyoto, Japan\\
National Institute for Basic Biology, Okazaki, Japan\\
lana.sinapayen@gmail.com} 

\begin{document}
\maketitle

\begin{abstract}
This paper reports on patterns exhibiting self-replication with spontaneous, inheritable mutations and exponential genetic drift in Neural Cellular Automata. Despite the models not being explicitly trained for mutation or inheritability, the descendant patterns exponentially drift away from ancestral patterns, even when the automaton is deterministic. While this is far from being the first instance of evolutionary dynamics in a cellular automaton, it is the first to do so by exploiting the power and convenience of Neural Cellular Automata, arguably increasing the space of variations and the opportunity for Open Ended Evolution.
\end{abstract}

\section{Data Sharing}

The experiments in this paper are executable online with no other requirements than access to a web browser; the data and the data analysis scripts are also open access.

The code used to generate these experimental results is available as an interactive Colab notebook at \url{https://github.com/LanaSina/NCA_self_replication}, as well at the R code used to generate figures and selected videos. The author apologizes in advance for non-optimal code and crimes against Tensorflow.

The data is available on Figshare: \url{https://figshare.com/projects/Self-replicating_Neural_Cellular_Automata/167582}

Videos are available at: \url{https://youtube.com/playlist?list=PLYuu1RcSnrYRhophmfolv_lmx7Qz8AP1P}

\section{Introduction}

Can a closed world, with unchanging rules and without outside influences, produce seemingly endless novelty?
This concept called ``Open Ended Evolution" is an unsolved problem in Artificial Life: the \textit{Evolution Prize for open-ended evolutionary innovation in a closed system} has been unclaimed for 17 years~(\cite{OEEprize}). 
The laws of physics in the time frame of the evolution of Life on Earth can be considered unchanged, but as pointed out in~\cite{OEEprize}, it is unclear if Earth can be considered a closed system, due to its known (and hypothetical) exchanges with the rest of the Universe. By contrast, the ``laws of biology" are sometimes considered to be continuously changing~(\cite{adams2017formal}), despite being implemented using unchanging physical laws. Is this a simple issue of definition, or a meaningful contradiction? In this paper, we propose to use Neural Cellular Automata (NCA,~\cite{mordvintsev2020growing}) to model closed worlds with unchanging rules, and find out how close we can get to Open Ended evolutionary dynamics. 
Cellular Automata are programs that run on a grid where each cell is defined by its state. The state of each cell is updated depending on this cell's previous state and the state of its neighbors, according to a fixed set of rules.

Cellular automata have been used to model some of the functions of Life from the very beginning of their invention. Von Neumann, in his search for a ``complicated artificial automata" of which complexity would grow under natural selection, used a cellular automaton to show that self-replication with inheritable mutation is possible in an artificial system~(\cite{neumann1966theory}). Conway named his most famous cellular automaton ``Life"~(\cite{izhikevich2015game}), and was interested in finding complex dynamics even before ``Life" was found to be Turing complete. In 2017, a biological von Neumann cellular automata was even found to be implemented on the back of a lizard~(\cite{manukyan2017living}). But most cellular automata have to be hand designed by the experimenter, who either implements known rules driving a given phenomenon (e.g. predator-prey systems~(\cite{cattaneo2006full}), reaction diffusion~\cite{weimar1997cellular}), or searches for rules that give an output similar to a known phenomenon (\cite{schepers1992two}). To facilitate this tedious design process, even before early NCA~(\cite{li2001calibration}), there was some interest in automating the discovery of relevant rules through optimization~(\cite{clarke1997self}).  The recent advances in deep learning may make this task easier, especially the open source, fast converging model proposed by~\cite{mordvintsev2020growing}. NCA only require the experimenter to define an initial state, a target state and a maximal number of computation steps. The rules that make the transformation possible are learned by the network, instead of being designed by the experimenter.
While they have not yet been used to model evolution, NCA have proven useful to implement biology-like functions in artificial patterns. Patterns that grow from a seed~(\cite{mordvintsev2020growing}), similar to the development of biological organisms from egg to adult form; patterns that self-repair~(\cite{mordvintsev2020growing, horibe2021regenerating}); patterns that undergo metamorphosis~(\cite{najarro2022hypernca}), or parasite and highjack other patterns~(\cite{randazzo2021adversarial}). Many non-neural cellular automata have explored the possibility of (Open Ended) evolution, with several organisms controlled by one automaton (\cite{sayama1999new, oros2007sexyloop}). Evoloops in particular, while limited in their phenotype diversity (square loops), show complex genetic evolutionary dynamics. The ``organisms" in~\cite{adams2017formal}, while highly abstracted from common definitions of organisms or evolution, are even used to formally define Unbounded Innovation and Unbounded Evolution. Yet most publications on NCA use a one-to-one mapping between automata and organisms: one organism is modeled by one automata, and if two organisms interact (for example through parasitism) each follows the rules of its own dedicated automaton. We know of 3 exceptions: \cite{otte2021generative}, where a NCA is trained to in-paint several images from edges, \cite{cavuoti2022adversarial}, where two set of rules are explicitly encoded in the model using hand-designed constraints; and \cite{hybrids} where several organisms are grown and hybridized. Note that these works are not about evolution and therefore do no not have self-replication mechanisms, but~\cite{hybrids} in particular (while not being a traditional publication) shows interesting developmental modularity.

In this paper we merge the world-rule approach of non-neural cellular automata and the convenience of NCA:  we consider each NCA as a world with rules loosely equivalent to the laws of physics of that world, and focus on the issue of self-replication, diversity of organisms, and evolution. In our experiments, we present training techniques that result in self-replication, spontaneous mutations, inheritance, and exponential genetic drift in NCA.

\section{Methods}

\subsection{Neural Cellular Automaton}

\begin{figure}[t]
\begin{center}
\includegraphics[width=\linewidth]{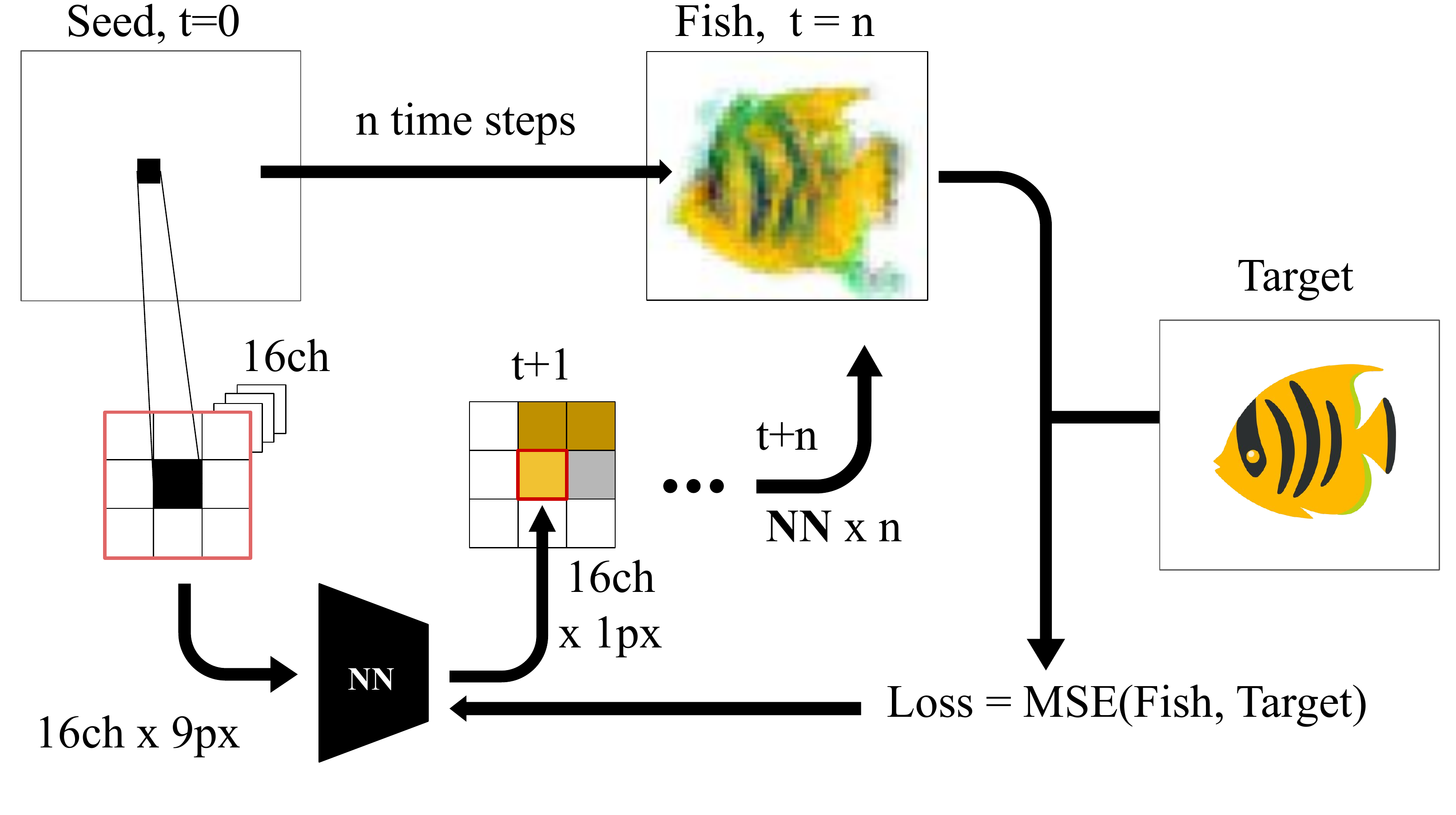}
\caption{\textbf{The NCA is trained to generate a target state in \textit{n} time steps starting from an initial seed state.} Each cell of the automaton's grid is represented by a pixel. A cell's state is a vector or 16 real numbers, 3 of which are used as RGB input to render the image, and 1 is the Alpha channel used to determine if a cell is alive or dead. The remaining 12 values are free parameters. A neural network (NN) takes the 9-cell neighborhood as input, and outputs the updated state of the central cell. The NN is applied to alive (Alpha$>$0.1) cells in the grid over \textit{n} time steps, then the loss is calculated between the RGB channels of this final state and the target state.}
\label{fig:nca}
\end{center}
\end{figure}

This project uses 2-dimensional Neural Cellular Automata~(\cite{mordvintsev2020growing}). Like many cellular automata, NCA run on a grid where each cell is defined by its state. This model is therefore fully spatially discrete, not continuous. The state of a cell is updated depending on the cell's previous state and the state of the cell's neighbors, according to a fixed set of rules. The main characteristic of a NCA is that these update rules are encoded by a Neural Network~(Noted \emph{NN} in Fig.~\ref{fig:nca}). In this paper, the state is a vector of 16 real values between -1 and 1, with the first 4 values corresponding to RGBA channels used to render an image on the NCA's grid. The NCA can therefore be trained using images as RGB targets that the grid must reach from its initial state. The Alpha channel determines whether a cell is alive (Alpha$>$0.1) or dead (Alpha$\le$0.1). If the cell is alive, the update rules apply: the cell's state is recalculated by applying the neural network to the neighborhood of the cell. If the cell is dead, the state vector is reset to 16 zeros and no update is applied. During training, the NCA's life span, i.e. one training step, is the number of updates (\textbf{time steps}) allowed to reach the target state from the initial state (typically 1~training step~=~96~time steps, in keeping with the original NCA paper, except when indicated otherwise). After one training step, the final state of the NCA is evaluated against the target image using the mean squared error as loss function. The weights of the network are updated through gradient descent.
Training ends when the maximum number of \textbf{training steps} has been reached (a number determined ad hoc by judging loss convergence). The model is then ready to be used and the rules do no change past this point.

\subsection{Modified training for self-replication}

\begin{figure}[t]
\begin{center}
\includegraphics[width=\linewidth]{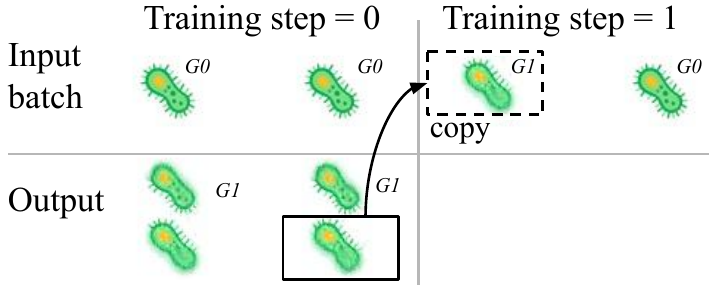}
\caption{\textbf{Batch substitution.} We replace half of the input batch at training step~1 with the output of the NCA at training step~0. This allows the NCA to learn to self-replicate its own output while simultaneously staying close to the target image. There are 8 batches but for simplicity only 2 are shown here.}
\label{fig:batches}
\end{center}
\end{figure}

Compared with the original NCA paper, we modify the training procedure of the NCA for some of the experiments:
(a) Batch substitution. Like most modern neural networks, the NCA is trained by batches: rather than one initial state, a batch of 8 copies of the initial state are updated at once. In the experiments with ``batch substitution", we replace half of the batch with the previous output of the NCA, as shown on Fig.~\ref{fig:batches}.
(b) Target alternation. In experiments with several target states instead of one target state, we alternate between the targets at each training step.
(c) Synchronous update rules. The original update rules are asynchronous: at each time step, half of the cells are chosen at random to be updated and the other half remain unchanged. In some experiments we instead use synchronous update rules (all cells are updated simultaneously). The training of the asynchronous models succeeds the vast majority of the time (i.e the loss converges), to the point that it is difficult to produce statistics on the failure rate. The synchronous models fail to converge much more often, however when they converge, the results are quantitatively similar to the asynchronous models, if producing qualitatively smoother images transitions.
(d) Periodic boundary conditions. The grid size is slightly more than twice the target pattern's size.
Other parameters are the same as~\cite{mordvintsev2020growing}, most importantly the neighborhood of radius 1 giving a neighborhood of 9 cells, the threshold of 0.1 on the Alpha channel to consider a cell alive rather than dead, and the 2-layer neural network to learn the update rules.

\subsection{Calculating the genetic drift}

This model has no alleles or chromosomes in the DNA representation, so our definition of genetic drift is different from the biological definition. We define genetic drift as the accumulation of neutral mutations in the genetic code through successive generations. In the absence of selection, all mutations in the model are neutral, except from the rare mutations that prevent an organism from replicating; the possibility of these mutations is largely eliminated during training and therefore rare after convergence of the model.

\begin{figure}[t]
\begin{center}
\includegraphics[width=\linewidth]{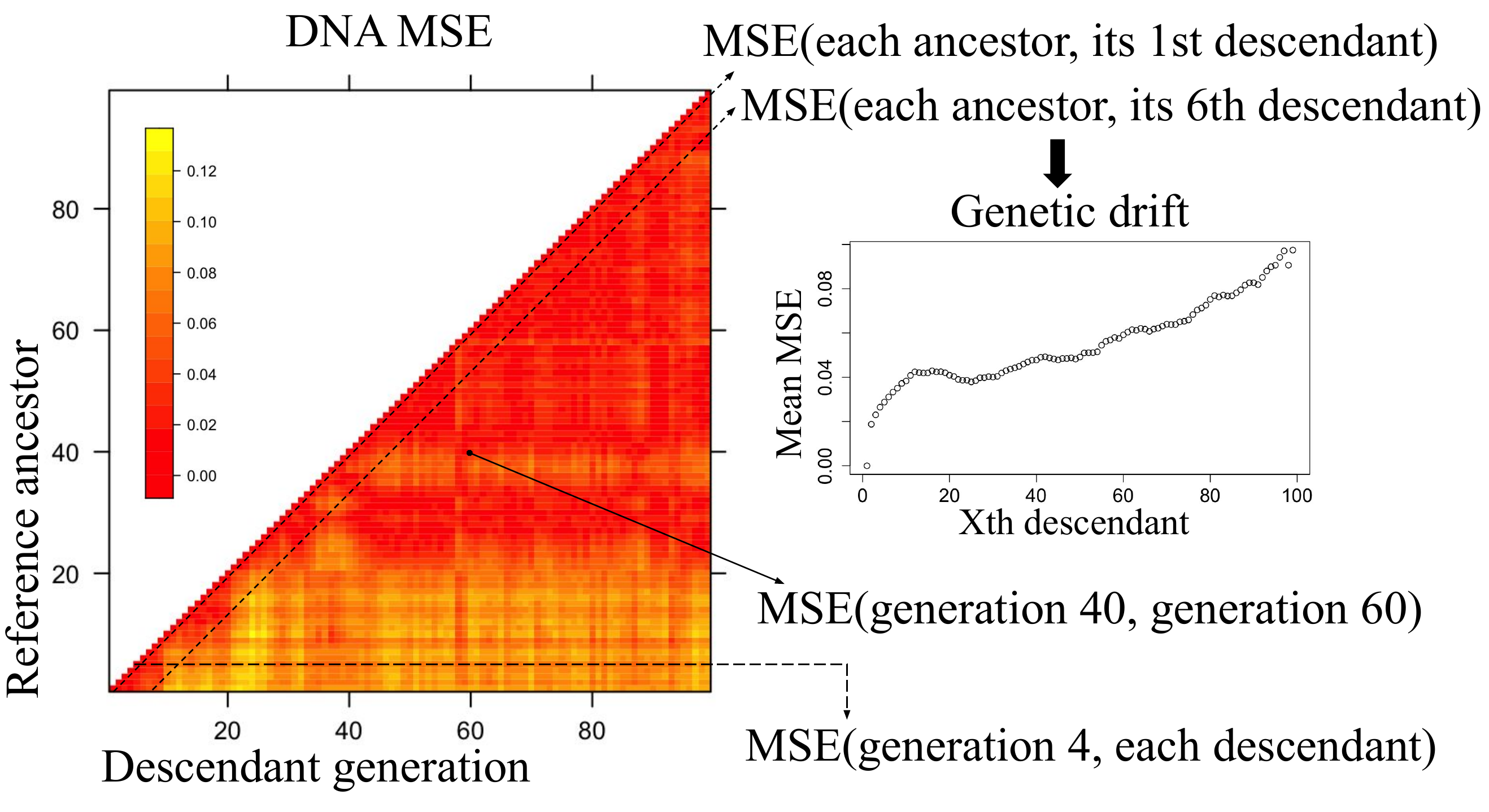}
\caption{\textbf{Genetic drift.} We calculate the Mean Squared Error~(MSE) between the DNA of an organism and each of its descendants. This value is represented by the color on the heatmap. One row represents the MSE of all generations relative to one reference ancestor: for example row 4 is the MSE of generations 5 to 100 relative to generation 4. Therefore, a diagonal on the heatmap represents the MSE of all pairs of [ancestor, Xth descendant]. Calculating the average value of each diagonal gives us the genetic drift through generations.}
\label{fig:heatmap}
\end{center}
\end{figure}

To calculate genetic drift, we use models were organisms have two clear life phases: growth and replication. Growth is the development of an egg (a small, square clump of black pixels) into a fully formed organism. Replication is the phase where an organism lays a new egg. We record the value of the state of all cells in the first egg laid by an organism, and call this value the DNA of the organism. The egg develops into an organism of its own, and this organism lays it own first egg. We record that DNA, and so on for 100 generations. Note that there is no fitness-dependant selection: the first offspring is always chosen. We calculate the Mean Squared Error (MSE) between the DNA of one organism and each of its descendants individually. This value is represented by the color on the heatmap of Fig.~\ref{fig:heatmap}.  One row on the heatmap represents the MSE of all generations relative to one reference ancestor: for example row 4 is the MSE of generations 5 to 100 relative to generation 4. Therefore, a diagonal of the heatmap represents the MSE of all pairs of [ancestor, Xth descendant]. For example,  the values on the 1st (longest) diagonal are the MSE between all parents and children. The second longest diagonal is the MSE between all grand-parents and grand-children, etc. So the average value of a diagonal is the average genetic distance between ancestor and Xth descendant.
Calculating the average genetic distance on an entire lineage gives us the genetic drift through generations: are an organism's grandchildren more genetically different from it than its children? Note that the number of data points decrease through time: for 100 generations, we have 99 pairs of parents and children, but we only have one pair of an organism and its 100th descendant.
The same method applied to the values of all cells of an adult organism (rather than just the egg) is used to calculate phenotypic drift.

\section{Results}

\subsection{Self-replication}

These experiments demonstrate a method to obtain patterns that self-replicate in a NCA. While most uses of NCA in the literature have one initial state A and a fully distinct target state B, we can instead train the NCA to go from A to 2A. After training, the model should be able to go from 2A to 4A, and so forth. In practice, the NCA becomes a replication function for ``exactly A", and any minute deviation A* from the target pattern stops the replication. Since the model is not pixel-perfect, its output is never exactly 2A, but rather 2A*, therefore replication always stops at this stage. The solution to this issue is to train the NCA to replicate anything ``close enough to A", by using batch substitution at each training step~(Fig.\ref{fig:batches}), replacing half of the batch of initial states A by the replicated states A* generated by the NCA itself. We use the bacteria emoji for this experiment, and the NCA learns to replicate its own output while simultaneously staying close to the target image, as shown in . While this training allows for deviations from the initial target, i.e. mutations, in theory there is no reason for these mutations to be inheritable or unbounded. (This is not in the scope of this paper, but it proved trivial to make the mutations non-inheritable.) In practice, in our small grid, the patterns rapidly crowd each other, so to investigate replication and mutations, we cut out individual patterns and transplant them to an empty grid.

\begin{figure}[t]
\begin{center}
\includegraphics[width=\linewidth]{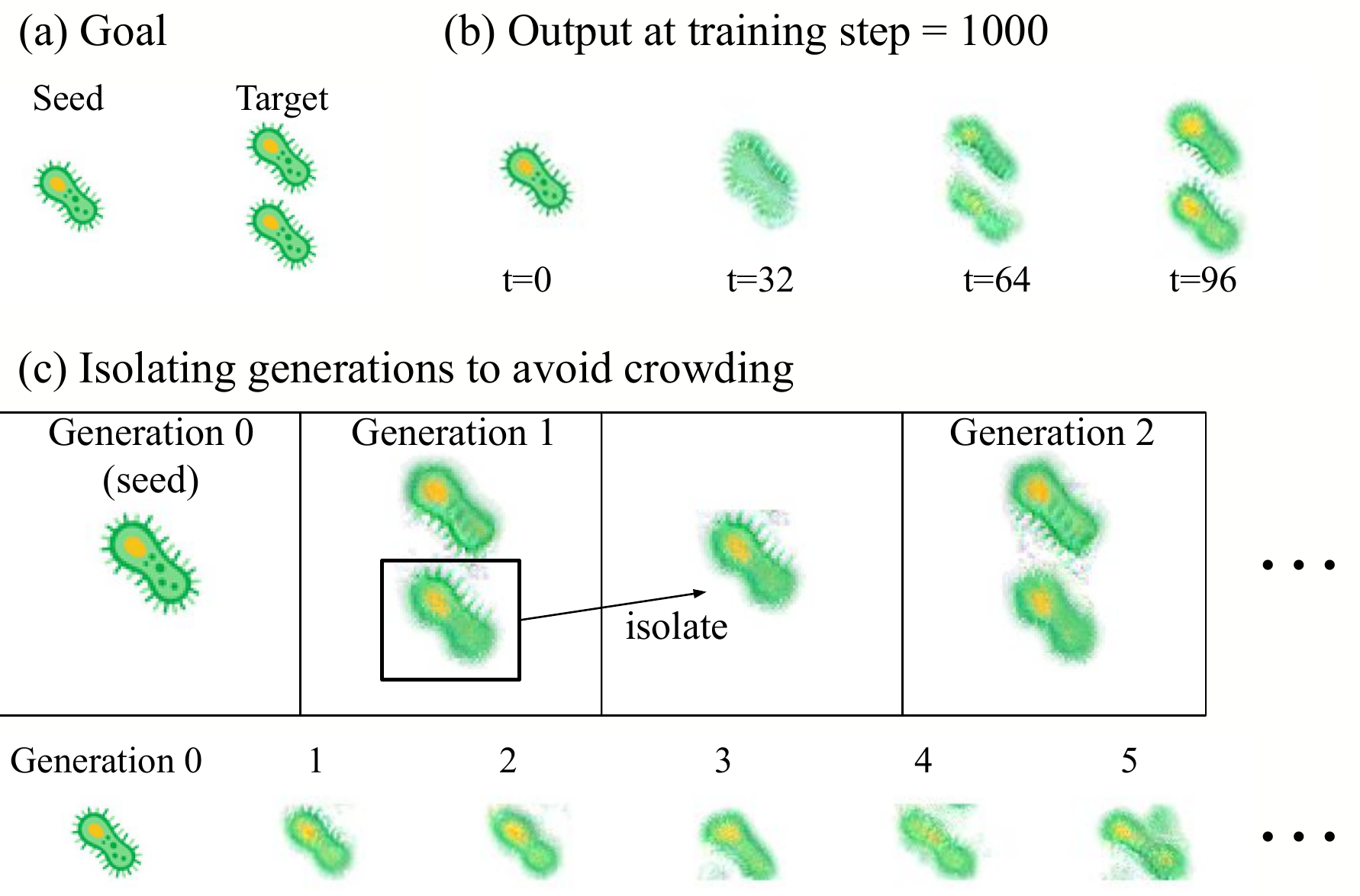}
\caption{\textbf{Simple self-replication.} (a) Results shown for a NCA is trained to self-replicate from a bacteria emoji. (b) To analyze the successive generations without interference from the grid becoming crowded with bacteria, we isolate one bacteria after replication (top or bottom, chosen randomly) and transplant it to a blank grid where it can replicate again. (c) Note the visual differences (mutations) between successive generations: G5 seems to have 2 nuclei (yellow central patch). Mutations extends to the non-RGBA values of each cell's state. Grid lines are not shown, but each pixel is a distinct cell of the automaton.}
\label{fig:bacteria}
\end{center}
\end{figure}

\begin{figure}[t]
\begin{center}
\includegraphics[width=\linewidth]{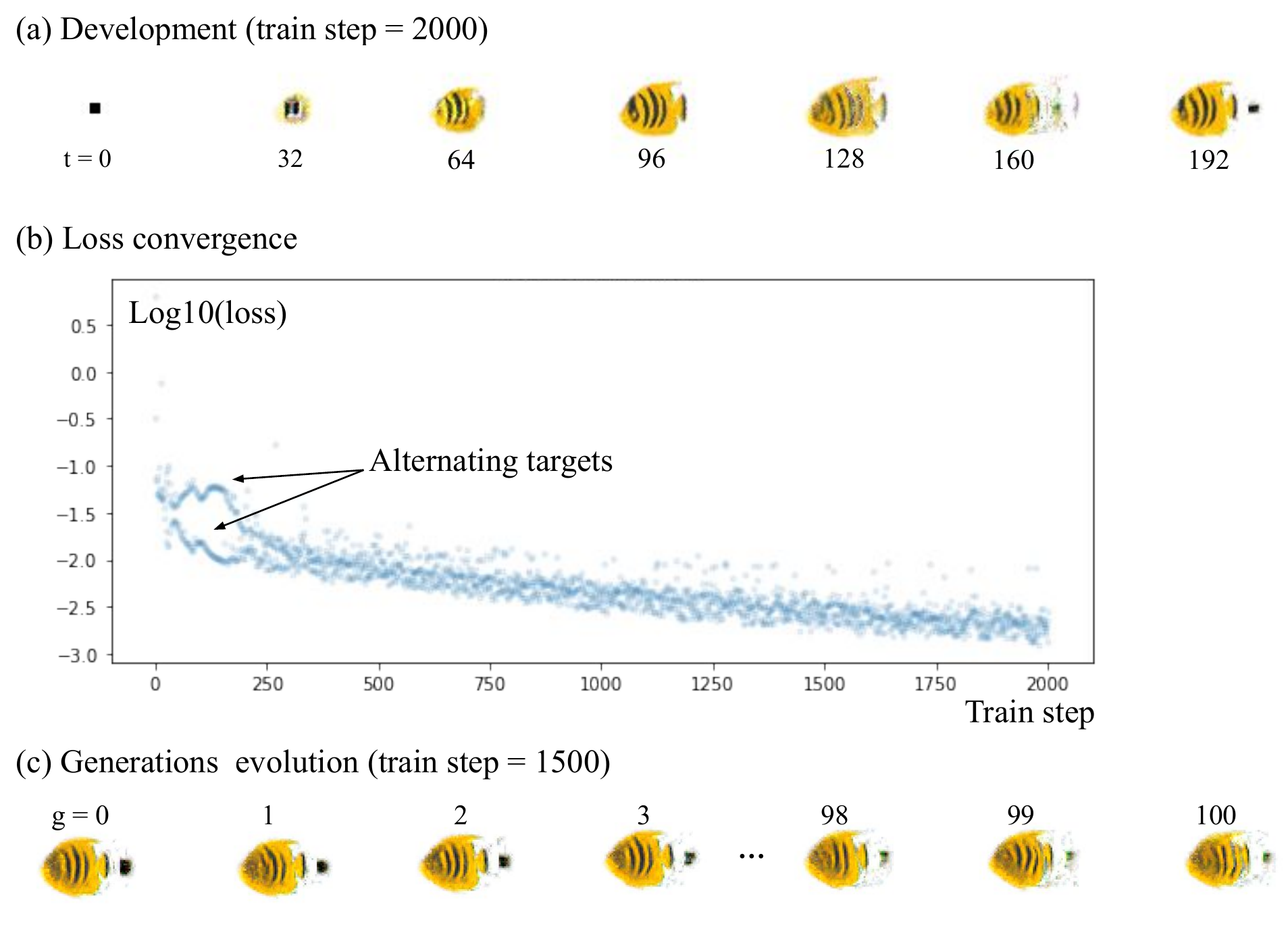}
\caption{\textbf{Growth and self-replication.} We add a self-replication step to the growth phase first introduced by~\cite{mordvintsev2020growing}. (a) The NCA is trained for 1000 training steps by alternating between an intermediary target (grow fish from egg) and a final target (move fish left and lay egg), as can be seen in the divided loss curve (b). (c) Successive generations show signs of mutation: by generation 98 the fish has lost one of the target's 3 stripes, but generation 99 regains it and generation 100 adds a supernumerary 4th stripe.}
\label{fig:fish}
\end{center}
\end{figure}

A more complex variation of self-replication is to have distinct growth and division phases, such as: A becomes B (growth), B becomes B+A (division). The two phases must be learned by the NCA using two different target states. We demonstrate through the following example, Fig.~\ref{fig:fish}: an egg grows into a fish (Target~1), the fish moves to the left and lays an egg (Target~2 includes both fish and egg). Compared to the bacteria experiment, this introduces one intermediary target, so during training we alternate between training the transition from egg to Target~1~(growth) and from Target~1 to Target~2 (division). Once again we use batch substitution for all training steps, and we transplant each egg to an empty grid to grow undisturbed. Results in ~Fig.~\ref{fig:fish} and in this \href{https://user-images.githubusercontent.com/18609788/224480411-7ba97be0-45ad-4013-9067-31b2df28ea19.mp4}{video link} show that we do obtain self-replication, and that successive generations show signs of mutation: by generation 98 the fish has lost one of the target’s 3 stripes, but generation 99 regains it and generation 100 adds a supernumerary 4th stripe.

\subsection{Spontaneous, inheritable mutations}

\begin{figure}[t]
\begin{center}
\includegraphics[width=\linewidth]{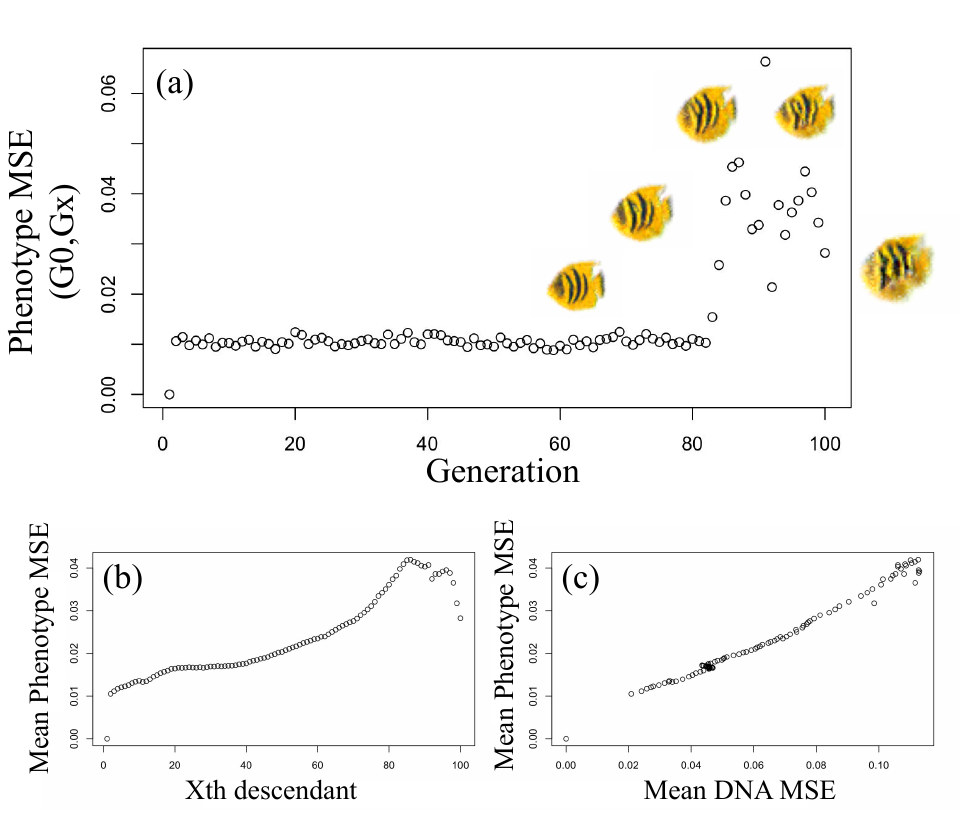}
\caption{\textbf{Genetic coding and drift.} A different run of the model in Fig.~\ref{fig:fish}, at training step 1500. (a)~MSE between the fish at generation 0 and its descendants. The descendants appear to be all equally different from the 0th generation, except for a jump at generation 82 where the fish develop a forked stripe that is inherited by successive generations. (b)~When calculating genetic drift, we find not a linear relationship as in~(a), but an exponential increase in MSE until generation 82, where this model stalls (not all models stall in 100 generations). (c)~The clear correlation indicates the emergence of a genetic code: DNA differences in the eggs are translated to phenotype differences in the developed organism, and big DNA mutations correspond (mostly) linearly to big phenotype differences.}
\label{fig:mutations}
\end{center}
\end{figure}

When comparing successive generations of fish, we can see that the offspring are always slightly different from the parents, suggesting that spontaneous mutations are occurring somewhere in the process tThe training process does not explicitly enforce any DNA-like coding or inheritance). By calculating the distance between the parents and offspring patterns, we find that there is indeed a form of inheritance, as mutations are carried from fish to egg and from egg to fish, therefore influencing the whole lineage.
Qualitatively, we see in Fig.~\ref{fig:fish}(c) a lineage where the 3rd black stripe of the fish was lost at generation 98, then gradually regained and followed by a 4th stripe. Fig.~\ref{fig:mutations}(a) shows a lineage where a mutation for a forked stripe develops over generations 80 to 90. Most mutations are not this obvious, and the more a model converges during learning, the less striking the mutations are. Quantitatively, Fig.~\ref{fig:mutations}(c) shows that DNA and phenotype are strongly correlated: a form of genetic coding has emerged in the model. Along with Fig.~\ref{fig:mutations}(b), it also shows genetic and phenotypic drift along generations, a topic we explore in the next section.

The main source of stochasticity in the NCA is the asynchronous update rule. The synchronous model's training is more brittle and often fails to converge, especially if the training has several targets. However, for successful training on the bacteria division task, we still find substantial inheritable mutations through generations (\href{https://user-images.githubusercontent.com/18609788/224480381-cd702897-b867-444e-bc51-b1577846e2bd.mp4}{video link}). These mutations despite the NCA rules being deterministic could be due to rounding errors that often occur with floating point number representation.
It is also possible that stochasticity is introduced elsewhere in the model unbeknownst to the experimenter, or due to the equipment (stochasticity in GPU runs). Note that these causes would still satisfy the definition of closed model by~\cite{OEEprize}. Finally, there is the possibility that each division is inherently different from the previous one, i.e. that the model that is genuinely deterministic but chaotic. This last hypothesis is reinforced by the fact that running the synchronous model from the same starting point always seems to lead to similar final results, even if those results are far from the initial state. This is not the case with asynchronous models, and we focus our analysis on those models in the remainder of the paper.

\subsection{Genetic encoding and exponential drift}

If each NCA is a world with its own laws of physics, what happens when we transplant a creature from one world to another? The transplanted pattern could disintegrate, maintain itself, or become a sort of hybrid.
We found that a NCA trained to develop a fish emoji from an egg will convert all input information into fish, including noise or other images, a disappointing but understandable result. Because of the training process, NCA have one overwhelming drive: to develop towards the target pattern. Unlike real worlds (and similarly to teleological misunderstandings of evolution on Earth...) they have a goal that they are trained to always converge towards. This might also be the cause for the sudden stalling of the ``near-exponential" curve of Fig.~\ref{fig:mutations}(b), although some instances of this model do not stall at 100 generations (because of time constraints, we did not perform a systematic analysis of stalling). It would make sense for the model to reach some limits given its limited expressivity: there are only so many yellow fish one can draw on 70 square pixels.

\begin{figure}[t]
\begin{center}
\includegraphics[width=\linewidth]{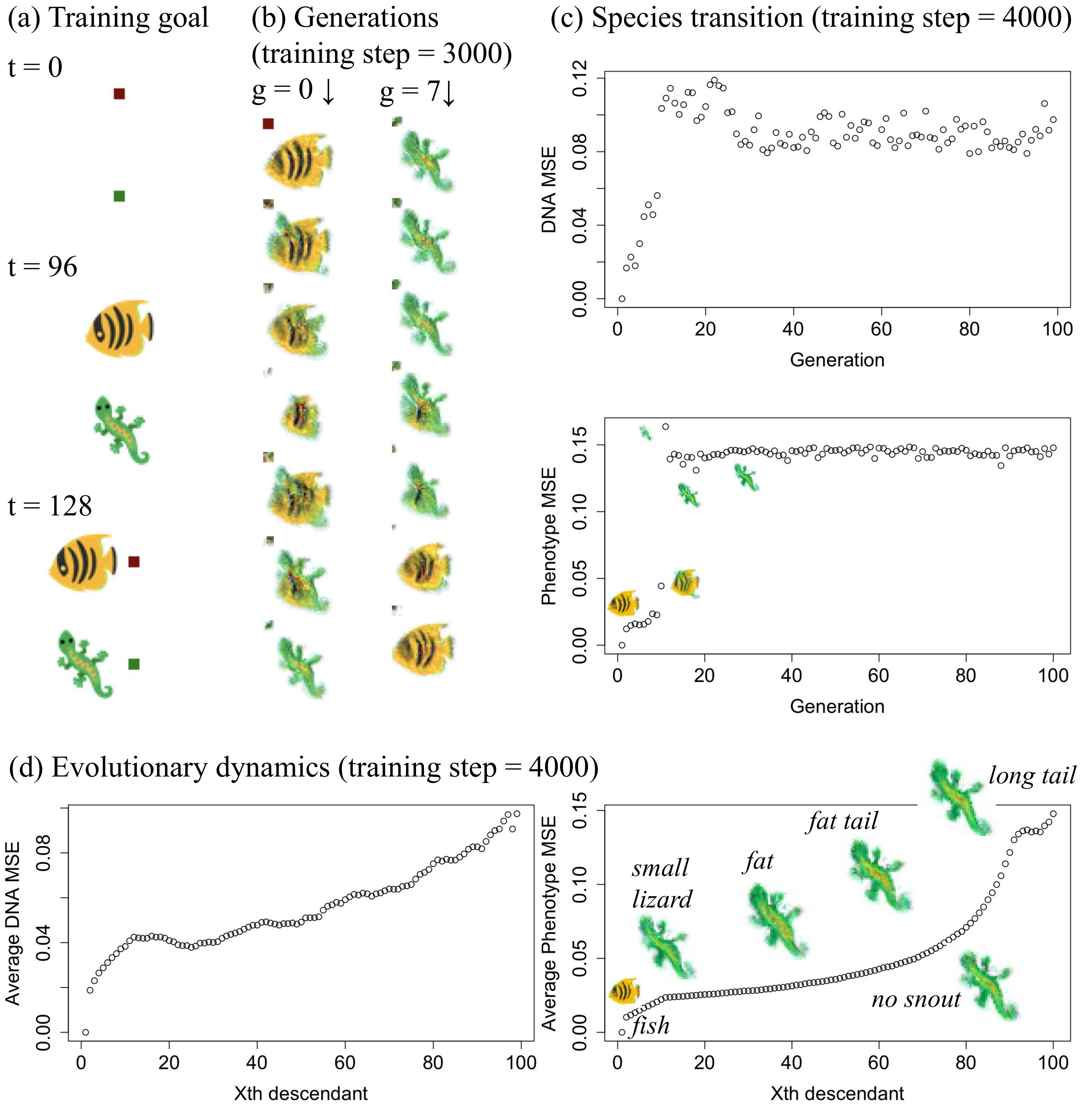}
\caption{\textbf{Exponential drift in a 2-organism NCA.} (a) The model is trained to develop different DNA into 2 different patterns, as well as replicating the patterns. (b) In some cases, especially before training convergence, the model goes back and forth between fish, lizard, and hybrids in the same lineage. (c) In this other lineage the DNA undergoes a relatively smooth transition, while the phenotype abruptly switches from fish to lizard. (d) The phenotype space is large, and the average difference of DNA (genetic drift) and phenotype (phenotypic drift) between an ancestor and its descendants increases exponentially.}
\label{fig:fish_lizard}
\end{center}
\end{figure}

A simple solution would be to train several NCA and execute several sets of rules within one space, but that would be a step back towards the concept of one NCA for one organism, and away from our goal of more than one organism in one self-contained NCA world.
One way to have several patterns coexist within a single set of rules is to train several eggs to converge to several targets. If the eggs contain the same information, the NCA converges to an average of the several targets, as the same rules apply to all cells of the NCA. If the eggs contain different information, similar to genetic information guiding development, we do obtain different patterns and occasionally stable hybrids of the patterns (Fig.~\ref{fig:fish_lizard}(b)). In addition, the exponential shape of the genetic drift curve is maintained.

Fig.~\ref{fig:fish_lizard}(c) to (d) show quantitative and qualitative analysis of 100 consecutive generations for a model trained for 4000 training steps. We note here that unlike models trained on single patterns, the lineages exhibit frequent extinction: some patterns are not viable and fail to produce eggs. The analysis was done on a run where 100 consecutive generations were viable. In some cases, especially in the early stages of training (e.g. training step = 3000, Fig.~\ref{fig:fish_lizard}(b)) the phenotypes switch with variable smoothness from one pattern to the other. This is less frequent as training progresses and the model converges. There is exponential drift of the descendants away from the ancestor (Fig.~\ref{fig:fish_lizard}(c)), and for the same magnitude of DNA variation (max.~0.10), the magnitude of phenotypic variation is higher: 0.15 for the fish-and-lizard model versus 0.04 for the fish only model. In other words, the same DNA-space codes a greater variety of phenotypes. Although we did not perform a quantitative analysis of when the exponential stalls, it is natural to expect that the qualitatively greater variety of phenotypes indicates a greater space of possibilities, and therefore longer or larger exponential growth than the 1-pattern model. The increase in lineage extinction events, while unexplained, is a caveat to this expectation.  All in all, the goal of creating several attractors and paths between them is achieved. 

However, this solution is still unsatisfying, as it could well be closer to ``paint by numbers" than to a modular genetic code. In the worst case, a fish's appearance could in theory be fully decided by one out of hundreds of bit being 0 in one egg, and all other values could lead to a lizard. The current training method does not guarantee that the NCA will not use DNA as a discrete identifier.

\begin{figure}[t]
\begin{center}
\includegraphics[width=\linewidth]{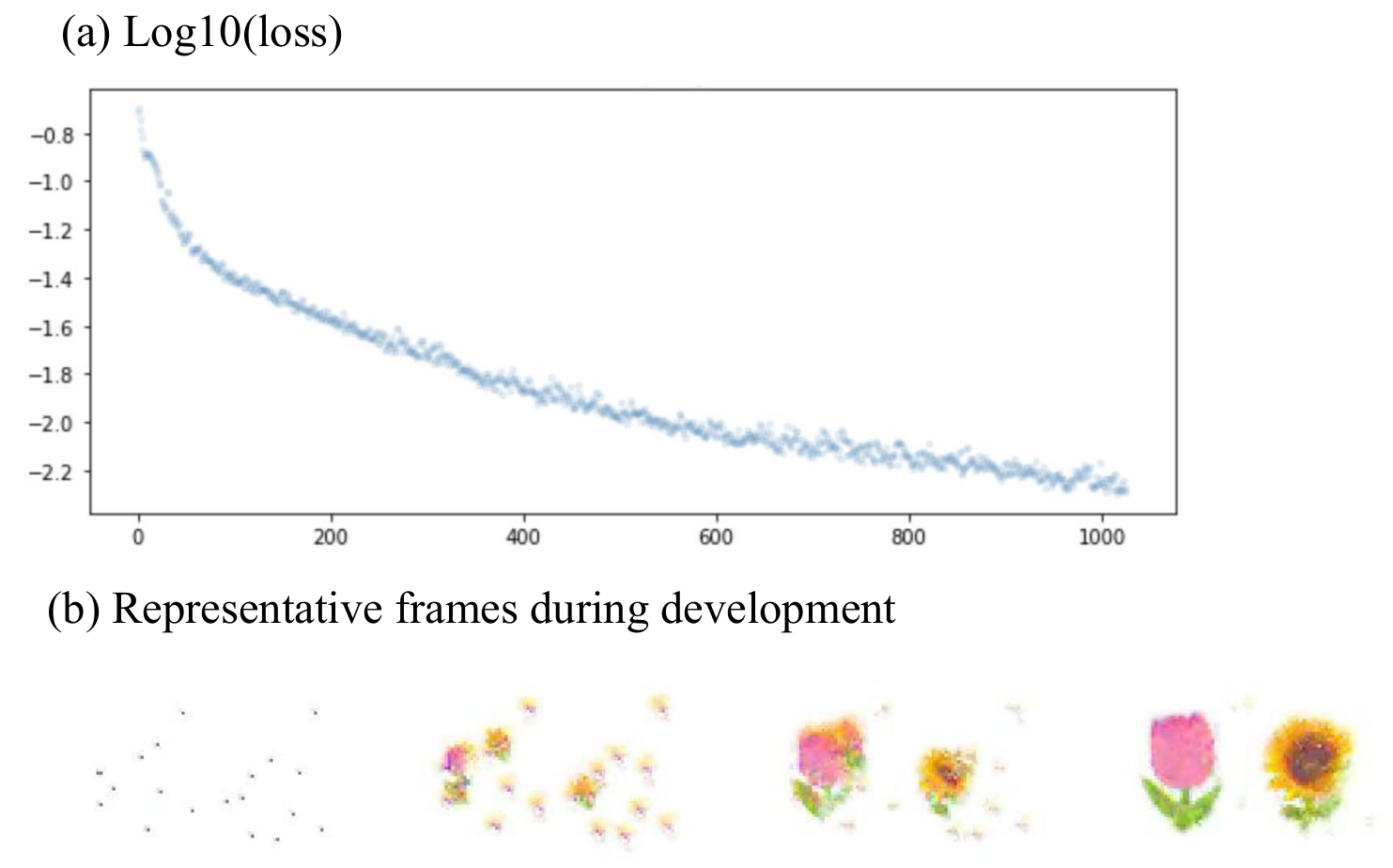}
\caption{\textbf{Spatial modular encoding:} (a) Smooth convergence of the spatial modular model. (b) In the initial state, all seed pixels (in black) have the same value, therefore they all develop the same patterns until they come into contact with each other. The directionality and timing of the contact is a cue for differentiation. Some undifferentiated patches differentiate into parts of the flowers and some disappear. The differentiation from initially identical instruction is reminiscent of the modularity of DNA.}
\label{fig:flowers}
\end{center}
\end{figure}

Other coding schemes are possible, for example Fig.~\ref{fig:flowers} shows the results of using a ``seed cloud" to code for different pattern. The scheme has some similarities with~\cite{hybrids}, except that all our seeds contain exactly the same genetic code. In consequence, the seeds initially develop identically, until they make contact with each other and the location of the contact serves to break the symmetry.  The directionality and timing of the contact is a cue for differentiation. Only the different positions of the seeds encode the final result. Since each seed starts with the same development but ends up being a different part of the final pattern(s), this might be closer to the type of modularity that characterises a DNA as we know it, where most cells of an organism have the same DNA and are undifferentiated until they are in the right neighborhood at the right time, at which point they differentiate into their final form.

\section{Discussion}

Using modified training methods, we show that NCA can exhibit self-replication and spontaneous, inheritable mutations, with runaway dynamics that carry the descendants' genetic code and phenotype away from their ancestors', even in the absence of selection. The expression of the mutations in the organisms' phenotypes are varied, non-repeating, and unexpectedly interesting, with stable inter-species mutants, addition or deletion of stripes in the fish experiment, doubling of the nucleus in the bacteria experiment, and various changes of size. While our experiments satisfy the definition of unbounded innovation and unbounded evolution by~\cite{adams2017formal} and arguably manages to implement changing biological rules as a subset of fixed physical laws, we still find weaknesses in the model. 1. There is no true unlimited diversity of organisms: a model trained to make lizards and fishes never grows a flower, even under directed evolution (where the experimenter imposes a fitness criterion to select organisms). 2. Due to ``crowding", the models presented here stop short of exhibiting actual human-out-of-the-loop evolution. While this paper does not discuss the notion of evolutionary complexity, and innovation is left undefined in the Open Ended Evolutionary Innovation prize~(\cite{OEEprize}), we feel that it might not even be warranted to talk about innovation in the absence of function, and our organisms have no functions related to their own survival besides ``lay an egg".

Ideally, selection would occur by itself and we could observe something closer to Open Endedness, where the basic laws of the world are largely fixed and yet life exponentially grows in complexity through real innovations. There are two major theoretical obstacles to this. Firstly, the organisms in NCA tend to suffer from crowding: because they are closer to waves of information than to physical matter, they can intersect each other and create information from nothing until the grid is ``full", rather than competing for space. A specific mechanism must be introduced for the patterns to have adversarial interactions.

Secondly and most importantly, the trained models lack expressivity. Deep Neural Networks are made for convergence, and by default NCA converge to one attractor: we must coax them to divergence, to obtain expressive power sufficient for several patterns to coexist. This might be possible by explicitly training the model for inheritable extraordinary mutations, which we have not done here, or by using a brute force approach and training one model on hundreds of target patterns, creating a hundred or more attractors.

This might be one of the big differences between AI, which strives for convergence, and ALife, which dreams of divergence. NCA being at the crossroads of both fields makes this conflict more salient. The limitations of NCA force us to imagine biologically implausible paths to evolution, another fundamental aspect of enjoying ALife research. While the mutations presented here are not adaptive, they do accumulate exponentially in the absence of evolutionary pressure, demonstrating perhaps the potential for true Open Ended Evolution in NCA. 

\footnotesize
\bibliographystyle{apalike}
\bibliography{example} 

\begin{thebibliography}{}

\bibitem[Adams et~al., 2017]{adams2017formal}
Adams, A., Zenil, H., Davies, P.~C., and Walker, S.~I. (2017).
\newblock Formal definitions of unbounded evolution and innovation reveal
  universal mechanisms for open-ended evolution in dynamical systems.
\newblock {\em Scientific reports}, 7(1):1--15.

\bibitem[Cattaneo et~al., 2006]{cattaneo2006full}
Cattaneo, G., Dennunzio, A., and Farina, F. (2006).
\newblock A full cellular automaton to simulate predator-prey systems.
\newblock In {\em Cellular Automata: 7th International Conference on Cellular
  Automata, for Research and Industry, ACRI 2006, Perpignan, France, September
  20-23, 2006. Proceedings 7}, pages 446--451. Springer.

\bibitem[Cavuoti et~al., 2022]{cavuoti2022adversarial}
Cavuoti, L., Sacco, F., Randazzo, E., and Levin, M. (2022).
\newblock Adversarial takeover of neural cellular automata.
\newblock In {\em ALIFE 2022: The 2022 Conference on Artificial Life}. MIT
  Press.

\bibitem[Cisneros et~al., 2022]{hybrids}
Cisneros, H., Sivic, J., and Mikolov, T. (2022).
\newblock Open-ended creation of hybrid creatures with neural cellular
  automata.
\newblock https://github.com/hugcis/hybrid-nca-evocraft.

\bibitem[Clarke et~al., 1997]{clarke1997self}
Clarke, K.~C., Hoppen, S., and Gaydos, L. (1997).
\newblock A self-modifying cellular automaton model of historical urbanization
  in the san francisco bay area.
\newblock {\em Environment and planning B: Planning and design},
  24(2):247--261.

\bibitem[Horibe et~al., 2021]{horibe2021regenerating}
Horibe, K., Walker, K., and Risi, S. (2021).
\newblock Regenerating soft robots through neural cellular automata.
\newblock In {\em Genetic Programming: 24th European Conference, EuroGP 2021,
  Held as Part of EvoStar 2021}, pages 36--50. Springer.

\bibitem[Izhikevich et~al., 2015]{izhikevich2015game}
Izhikevich, E.~M., Conway, J.~H., and Seth, A. (2015).
\newblock Game of life.
\newblock {\em Scholarpedia}, 10(6):1816.

\bibitem[Klyce, 2006]{OEEprize}
Klyce, B. (2006).
\newblock The evolution prize: Is open-ended evolutionary innovation in a
  closed system possible?
\newblock https://www.panspermia.org/evolutionprize/.

\bibitem[Li and Yeh, 2001]{li2001calibration}
Li, X. and Yeh, A. G.-O. (2001).
\newblock Calibration of cellular automata by using neural networks for the
  simulation of complex urban systems.
\newblock {\em Environment and Planning A}, 33(8):1445--1462.

\bibitem[Manukyan et~al., 2017]{manukyan2017living}
Manukyan, L., Montandon, S.~A., Fofonjka, A., Smirnov, S., and Milinkovitch,
  M.~C. (2017).
\newblock A living mesoscopic cellular automaton made of skin scales.
\newblock {\em Nature}, 544(7649):173--179.

\bibitem[Mordvintsev et~al., 2020]{mordvintsev2020growing}
Mordvintsev, A., Randazzo, E., Niklasson, E., and Levin, M. (2020).
\newblock Growing neural cellular automata.
\newblock {\em Distill}.
\newblock https://distill.pub/2020/growing-ca.

\bibitem[Najarro et~al., 2022]{najarro2022hypernca}
Najarro, E., Sudhakaran, S., Glanois, C., and Risi, S. (2022).
\newblock Hypernca: Growing developmental networks with neural cellular
  automata.
\newblock {\em arXiv preprint arXiv:2204.11674}.

\bibitem[Neumann, 1966]{neumann1966theory}
Neumann, J.~v. (1966).
\newblock Theory of self-reproducing automata.
\newblock {\em Mathematics of Computation}, 21:745.

\bibitem[Oros and Nehaniv, 2007]{oros2007sexyloop}
Oros, N. and Nehaniv, C.~L. (2007).
\newblock Sexyloop: Self-reproduction, evolution and sex in cellular automata.
\newblock In {\em 2007 IEEE Symposium on Artificial Life}, pages 130--138.
  IEEE.

\bibitem[Otte et~al., 2021]{otte2021generative}
Otte, M., Delfosse, Q., Czech, J., and Kersting, K. (2021).
\newblock Generative adversarial neural cellular automata.
\newblock {\em arXiv preprint arXiv:2108.04328}.

\bibitem[Randazzo et~al., 2021]{randazzo2021adversarial}
Randazzo, E., Mordvintsev, A., Niklasson, E., and Levin, M. (2021).
\newblock Adversarial reprogramming of neural cellular automata.
\newblock {\em Distill}.
\newblock https://distill.pub/selforg/2021/adversarial.

\bibitem[Sayama, 1999]{sayama1999new}
Sayama, H. (1999).
\newblock A new structurally dissolvable self-reproducing loop evolving in a
  simple cellular automata space.
\newblock {\em Artificial Life}, 5(4):343--365.

\bibitem[Schepers and Markus, 1992]{schepers1992two}
Schepers, H.~E. and Markus, M. (1992).
\newblock Two types of performance of an isotropic cellular automaton:
  stationary (turing) patterns and spiral waves.
\newblock {\em Physica A: Statistical Mechanics and its Applications},
  188(1-3):337--343.

\bibitem[Weimar, 1997]{weimar1997cellular}
Weimar, J.~R. (1997).
\newblock Cellular automata for reaction-diffusion systems.
\newblock {\em Parallel computing}, 23(11):1699--1715.

\end{thebibliography}

\end{document}